\documentclass{article}
\usepackage{spconf,amsmath,graphicx}
\usepackage{booktabs}
\usepackage{multirow}
\usepackage{algorithm}
\usepackage{algpseudocode}
\usepackage{amssymb}
\usepackage{hyperref}

\title{Anatomical Attention Alignment Representation for Radiology Report Generation}
\name{Quang Vinh Nguyen $^{\dagger}$, Minh Duc Nguyen $^{\dagger}$\thanks{\textsuperscript{\textdagger}Equal contribution as co-first authors.}, Thanh Hoang Son Vo, Hyung-Jeong Yang, Soo-Hyung Kim$^{ \star}$\thanks{$^{\star}$Corresponding author:shkim@jnu.ac.kr}}

\address{Department of Artificial Intelligence Convergence, Chonnam National University, South Korea
}

\begin{document}
%
\maketitle
\begin{abstract}
Automated Radiology report generation (RRG) aims at producing detailed descriptions of medical images, reducing radiologists' workload and improving access to high-quality diagnostic services. Existing encoder-decoder models only rely on visual features extracted from raw input images, which can limit the understanding of spatial structures and semantic relationships, often resulting in suboptimal text generation. To address this, we propose Anatomical Attention Alignment Network (A3Net), a framework that enhance visual-textual understanding by constructing hyper-visual representations. Our approach integrates a knowledge dictionary of anatomical structures with patch-level visual features, enabling the model to effectively associate image regions with their corresponding anatomical entities. This structured representation improves semantic reasoning, interpretability, and cross-modal alignment, ultimately enhancing the accuracy and clinical relevance of generated reports. Experimental results on IU X-Ray and MIMIC-CXR datasets demonstrate that A3Net significantly improves both visual perception and text generation quality. Our code is available at \href{https://github.com/Vinh-AI/A3Net}{GitHub}.
\end{abstract}
\section{Introduction}
\label{sec:intro}

The automated generation of clinical chest radiograph reports has emerged as a transformative advancement. It seamlessly integrates medical imaging with diagnostic processes by providing radiologists with an intelligent tool to efficiently produce detailed, accurate, and clinically relevant reports, enhancing diagnostic workflows and supporting evidence-based decision-making.

Unlike conventional image captioning tasks \cite{xu2015show, Rennie2016SelfCriticalST, lu2017knowing}, which primarily aim to produce concise summaries of image content, radiology report generation requires generating comprehensive and structured textual descriptions that detail both normal and pathological findings. This task presents unique challenges due to the complexity of medical images, the need for precise alignment between visual and textual features. To address these challenges, various methodologies have been explored \cite{chen-etal-2020-generating, wang2022medical, nazarov2022importance}. For example, hierarchically structured LSTM networks \cite{yin2019automatic} and memory-driven modules \cite{chen-etal-2020-generating, chen-etal-2021-cross-modal} have been employed to enhance long-term memory retention and improve the generation of detailed descriptions. Another significant challenge is data deviation, where normal findings dominate the dataset, potentially overshadowing pathological cases. To mitigate this issue, researchers have explored strategies such as refining image-text attention mechanisms, aligning visual and textual features, and integrating external domain knowledge. Specifically, the study in \cite{liu-etal-2021-competence} requires supervision signals to consider leveraging the relationship between the two modalities (radiographs and reports). Meanwhile, \cite{chen-etal-2021-cross-modal, liu-etal-2021-competence} conduct representation weighting for different modalities to balance the contribution from either visual or textual guidance. Despite these advancements, accurate cross-modal alignment remains a major challenge due to the lack of annotated data for supervised learning. This limitation makes it difficult for models to effectively map textual descriptions to corresponding visual regions. A promising direction is leveraging structured visual representations, particularly anatomical region-based modeling, to enhance spatial awareness and improve semantic understanding.

In this paper, we propose a novel anatomical attention alignment (A3Net) framework that enhances visual-textual understanding by generating hyper-visual feature representations. We construct hyper-visual representations by integrating localized patch-level features with an anatomical entity dictionary, allowing the model to explicitly recognize and interpret the spatial and semantic attributes of anatomical structures. The constructed region-aware representation enhances the alignment of image and text features while also providing a direct learning objective for improving report generation. In summarize, our main contributions are:
\begin{itemize}
    \item We propose a structured dictionary of visual entities along with localized patch features to create hyper-visual representations that are explicitly aware of the semantic and spatial characteristics of each anatomical entity. These structured features improve spatial and semantic awareness, leading to more interpretable and accurate report generation.
    \item  We develop a region-aware learning framework that optimally aligns visual and textual features, allowing the model to associate image patches with corresponding textual descriptions more effectively.
    \item Extensive experiments on IU X-ray and MIMIC-CXR datasets demonstrate that our model achieves state-of-the-art performance in RRG task. 

\end{itemize}

\section{Related work}
\label{sec:related}
The encoder-decoder approach, widely recognized as the most successful framework for image captioning, has also become the predominant approach in radiology report generation \cite{chen-etal-2020-generating, chen-etal-2021-cross-modal,liu-etal-2021-competence,nazarov2022importance,wang2022medical,yin2019automatic,sun2024r2gen,you-etal-2022-jpg}. Most methods employing this architecture utilize CNN-based visual extractor as the image encoder \cite{chen-etal-2020-generating, chen-etal-2021-cross-modal}, where an input image is divided into a sequence of image patches. Subsequently, sequential decoders generate the target textual sequence based on the aggregated features from all patches.

Global visual feature is the most widely used feature type in existing studies as it utilizes raw visual features extracted from input images. However, a significant limitation of these methods is the substantial loss of region-specific information, which hinders the model's ability to effectively learn the characteristics of pathological changes within specific regions and leverage the strong relationships between different anatomical structures. This shortfall underscores the need for more region-aware approaches to improve the accuracy and relevance of generated radiology reports. To this line, our work is designed to enhance the model's perceptual capabilities by providing a more structured and semantically rich representation of the input.

\section{Method}
\begin{figure}[t!]
    \centering
    \hspace{-4mm}
    \includegraphics[width=0.5\textwidth]{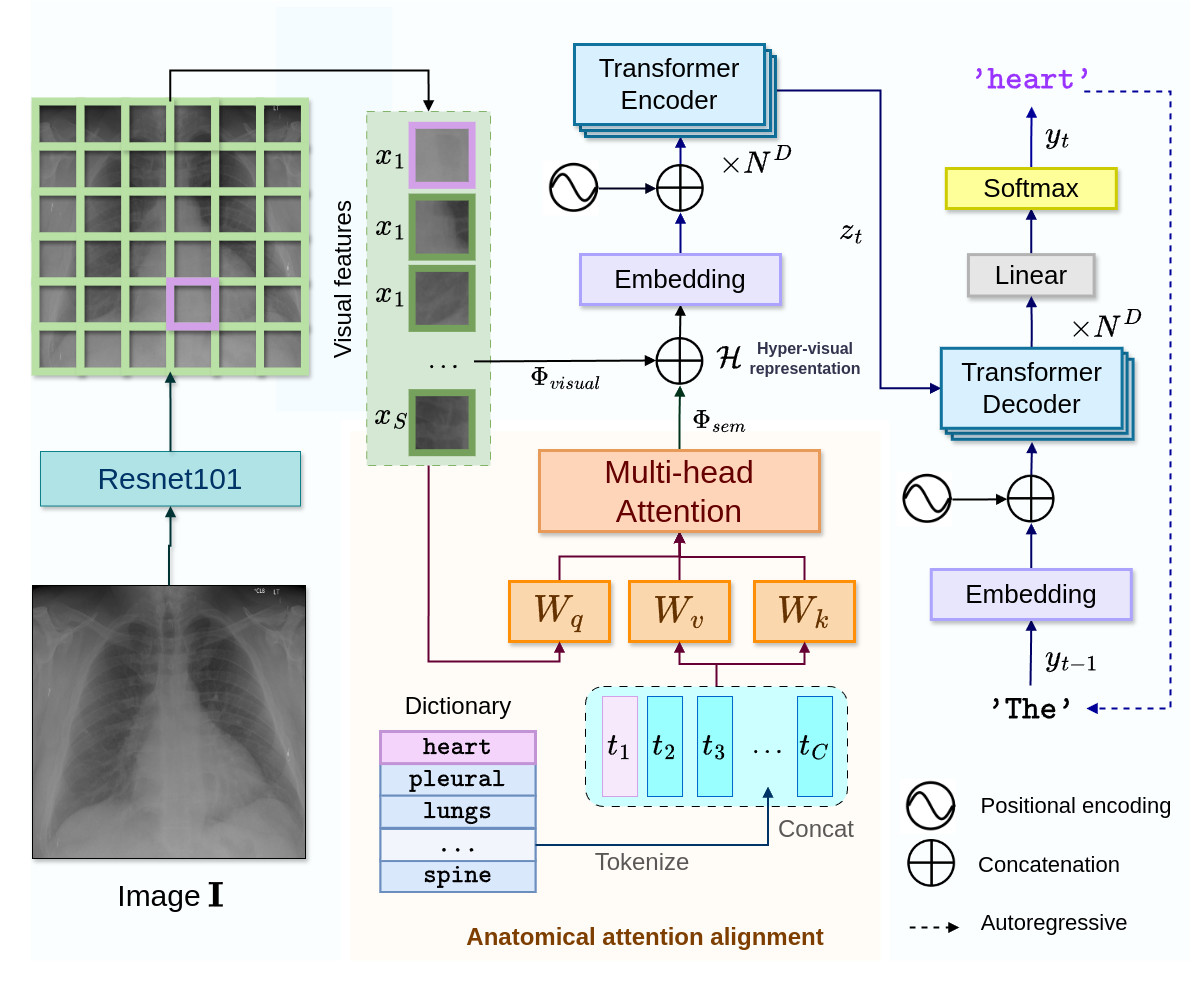} 
    \caption{
      The overall architecture of the proposed framework, which consists of a visual extractor, the anatomical attention alignment is highlighted in the light orange area, and a Transformer-based encoder-decoder for textual report generation.}
    \label{fig:overview}
\end{figure}
\subsection{Overview}
The objective of automatic chest X-ray report generation is to produce a coherent report as a sequence of generated tokens from chest X-ray images. In particular, we first extract input feature sequence of $S$ patches ${\mathcal{F}}_{\phi_\text{patch}}=\{x_1, x_2,...,x_S\}$ given a radiograph image $\mathbf{I} \in \mathbb{R}^{H \times W \times F}$ ($H, W, F$ are height, width, and view of the raw image, respectively), where $x_S \in \mathrm{R}^d$, then produce the output sequence of $T$ generated tokens $\mathbf{Y}=\{y_1, y_2,...,y_T\}$ with $y_i \in \mathbb{V}$ that describes and interprets the findings observed. Our model consists of a visual extractor $\mathcal{E}_{\phi_{I}}$, the anatomical attention alignment $\mathcal{A}_{\phi_{patch}}$ and $\mathcal{G}_{\phi_{\mathcal{H}}}$ is a captioning generator with source embedding parameters $\phi_{\mathbf{I}}$, $\phi_{patch}$, $\phi_{H}$ respectively. In brief, the target can be: 
\begin{equation}
    \mathbf{Y} =  \mathcal{G}_{\phi_{\mathcal{H}}} \circ(\mathcal{E}_{\phi_{I}}\oplus \mathcal{A}_{\phi_{patch}})
\end{equation}

The generation process is formulated as a recursive application of the chain rule, where the probability of the target sequence $\{y_1, y_2, \dots, y_T\}$ given the input image $Img$ is defined as:
\begin{equation}
p(\mathbf{Y}\mid \mathbf{I}) = \prod_{t=1}^{T} p(y_t \mid y_1, \dots, y_{t-1}, \mathbf{I}).
\end{equation}
The training objective involves maximizing the likelihood of the target sequence conditioned on the input image:
\begin{equation}
\theta^{*} = \underset{\theta}{\operatorname{argmax}} \sum_{t=1}^{T} \log p(y_t \mid y_1, \dots, y_{t-1}, \mathbf{I}; \theta), 
\end{equation}
where $\theta^{*}$ denotes the optimal parameters of the model. During inference, beam search is employed to generate predictions.
An overview of our proposed approach is presented in Fig.~\ref{fig:overview}. 
\label{sec:method}

\subsection{Anatomical attention alignment}

To generate radiology reports, we begin visual feature extraction from radiology images using Resnet backbone. As shown in Fig.~\ref{fig:overview}, the image is processed through $\mathcal{E}_{\phi_{I}}$ to produce a feature map. Each spatial location in the feature map corresponds to a specific region in the original image. These spatial features are then flattened into a sequential representation, which denoted as $\phi_{patch}$. This process can be formally expressed as: 

\begin{equation}
    \{ x_1 , x_2 , … , x_s \} = \mathcal{E}_{\phi_{\mathbf{I}}}(\mathbf{I})
\end{equation}

 Regions-aware Radiology Report Generation involves a critical aspect: the identification and interpretation of radiographic appearance-related attributes, such as anatomical structures (e.g., heart, lungs, ribs, diaphragm), organ morphology, and tissue texture. These attributes inherently demand a detailed spatial analysis to achieve accurate and meaningful text generation. Thus, we create hyper-visual representation to enrich the feature embedding:


\[
\mathcal{H} = \Phi_{\text{fusion}} \Big( \Phi_{\text{visual}} (\mathcal{E}_{\phi_{\mathbf{I}}}), \Phi_{\text{sem}} (\text{[Dictionary]}) \Big)
\]
where $\mathcal{H}$ denotes hyper-visual parameters. Our anatomical attention alignment strategy begins with diving visual features into patches. The initiation of the anatomical structures dictionary by pre-defining entity lists e.g., \texttt{\{pneumothorax, pleural, spine, heart, hernia,...\}}. Next, we apply token embedding to all anatomical entities and project them into a unified feature space. The anatomical attention alignment adopted semantic features and front-view features which can be indicated as $\mathcal{F_{\phi_\text{text}}}$ and ${\mathcal{F}}_{\phi_\text{patch}}$ are aligned by cross-attention. We use ${\mathcal{F}}_{\phi_\text{patch}} \gets \mathcal{E}_{\phi_{\mathbf{I}}}(\mathbf{I})$ as querry $Q$ and ${\mathcal{F}}_{\phi_\text{patch}}$ as a set of key-value pairs $K, V$ then adopted multi-head attention ($MHA$) as the as the core structure. The output is computed as a weighted sum of the values, with the weights determined by the dot-product similarity between the semantic and all visual features before being incorporated to enhance and enrich the visual features: 
\begin{equation}
\begin{aligned}
MHA(Q, K, V) &= \text{Concat} \Big( \text{softmax} \Big( \frac{Q W_i^Q (K W_i^K)^T}{\sqrt{d_k}} \Big) \\
&\quad \times (V W_i^V) \Big) W^O
\end{aligned}
\end{equation}

where \( W_i^Q, W_i^K, W_i^V \) are the learned projection matrices for each attention head \( i \),  
\( d_k \) is the dimension of the key vectors, softmax is applied row-wise, and \( W^O \) is the output projection matrix. We outline our anatomical attention alignment strategy in Algorithm~\ref{alg}.
\begin{algorithm}[ht!]
    \caption{Anatomical attention alignment}
    \footnotesize 
    \begin{algorithmic}[1]
        \State \textbf{Require:} \\
        \quad $\mathbf{E} = \{ e_1, e_2, \dots, e_C \}, \; e \in S$  \Comment{Dictionary with $C$ categories}  \\
          \quad ${\mathcal{F}}_{\phi_\text{patch}} \gets \mathcal{E}_{\phi_{\mathbf{I}}}(\mathbf{I}), \; {\mathcal{F}}_{\phi_\text{patch}} \in \mathbb{R}^{B \times (H \times W) \times F}$  \Comment{Visual features} 
       
\State \textbf{Process text features:}  \\
$\quad {\Phi_\text{text}} \gets \{ \mathbf{t_i} \mid \mathbf{t_i} = \text{tokenize}(e), \forall e \in \mathbf{E} \}$ \\
$\quad \mathcal{F_{\phi_\text{text}}} \in \mathbb{R}^{N \times D} \gets \bigcup_{i=1}^{N} \mathbf{t_i} \; \text{where} \; N = \text{len} ({\Phi_\text{text}})$ \Comment{Concat tokenized elements to form anatomical features}\\
        \quad $\mathcal{F_{\phi_\text{text}}} \gets \mathcal{F_{\phi_\text{text}}} \otimes 1_{Batch\_size}$ \Comment{Expand $\mathcal{F_{\phi_\text{text}}}$ to \( \mathbb{R}^{B \times N \times D} \)} 

        


\State \textbf{Cross-attention update:} \\
\quad \( Q = {\mathcal{F}}_{\phi_\text{patch}} \),  
\quad \( K = V = \mathcal{F_{\phi_\text{text}}} \) \\
\quad \( Q \leftarrow \text{LayerNorm}(Q + MHA(Q, K, V)) \)  \\
\quad \(\mathcal{F_{\phi_\text{sem}}} \leftarrow \text{LayerNorm}(Q + \text{FeedForward}(Q)) \)  \Comment{Semantic features}


        \State \textbf{Return} ${F}_{\phi_\text{patch}} \oplus \mathcal{F_{\phi_\text{sem}}}$  \Comment{Generate hyper-visual features}
    \end{algorithmic}
    \label{alg}
\end{algorithm}

\subsection{Captioning Generator}
Our captioning generator is built upon Transformer-based \cite{vaswani2017attention} encoder-decoder architecture. Provided the input sequence hyper-visual parameters $\mathcal{H}=\{h_1, h_2,..., h_S\}$, We first employ the standard Transformer encoder $f_e$ to obtain  the output sequence $\mathbf{Z}$ by $\{z_1, z_2,..., z_S\} = f_e(h_1, h_2,..., h_S)$.
Subsequently, we have integrated the Transformer decoder as the core component of our captioning generator. The decoding process follows an auto-regressive approach, expressed as:  $y_t = f_d(z_1, z_2, \dots, z_S, y_1, \dots, y_{t-1})$ where \( f_d(\cdot) \) represents the Transformer decoder.

\section{Experiments }
\label{sec:typestyle}
\subsection{Datasets and Metrics}
We perform our experiments on two widely used RRG datasets: IU-Xray \cite{demner2016preparing} and MIMIC-CXR \cite{johnson2019mimic}. In line with the experimental setups of prior studies \cite{li2018hybrid, chen-etal-2020-generating}, we focus on generating the findings section of the reports and exclude samples that lack this section for both datasets. The detailed statistics of the datasets are provided in Table~\ref{tab:dataset_stats}. We adopted a widespread combination of conventional natural language generation (NLG) metrics including BLEU, METEOR, and ROUGE-L to measure how well the generated reports match the ground truth reports. 
\begin{table}[b]
    \centering
    \footnotesize
    \renewcommand{\arraystretch}{1.2}
    \begin{tabular}{l ccc ccc}
        \toprule
        \textbf{Dataset} & \multicolumn{3}{c}{\textbf{IU X-Ray}} & \multicolumn{3}{c}{\textbf{MIMIC-CXR}} \\
        \cmidrule(lr){2-4} \cmidrule(lr){5-7}
        & \textbf{Train} & \textbf{Val} & \textbf{Test} & \textbf{Train} & \textbf{Val} & \textbf{Test} \\
        \midrule
        \textbf{Image }  & 5,226 & 748  & 1,496 & 368,960 & 2,991 & 5,159 \\
        \textbf{Report }  & 2,770 & 395  & 790 & 222,758 & 1,808 & 3,269 \\
        \textbf{Patient } & 2,770 & 395  & 790 & 64,586  & 500   & 293   \\
        \textbf{Avg. Len.}  & 37.56 & 36.78 & 33.62 & 53.00 & 53.05 & 66.40 \\
        \bottomrule
    \end{tabular}%

    \caption{Dataset statistics for IU X-Ray and MIMIC-CXR by their training, validation, and test splits, including image, report, and patient counts, along with the average report length (Avg. Len.).}
    \label{tab:dataset_stats}
\end{table}
\begin{table*}[t]
    \centering
    \renewcommand{\arraystretch}{1.2}
    \setlength{\tabcolsep}{5pt}
    \caption{Quantitative results of our proposed model with previous studies on IU X-RAY and MIMIC-CXR. The best values are highlighted in bold. BL, MTR, RG-L stand for BLEU, METEOR, ROUGE, respectively.}
    \label{tab:comparison}
    \begin{tabular}{l | l | c c c c c c}
        \hline
        \textbf{Data} & \textbf{Model} & \textbf{BL-1} & \textbf{BL-2} & \textbf{BL-3} & \textbf{BL-4} & \textbf{MTR} & \textbf{RG-L} \\
        \hline
        \hline
        \multirow{7}{*}{\textbf{IU X-RAY}} 
        &Show-tell \cite{xu2015show}& 0.308& 0.190& 0.125& 0.088& 0.256& 0.122 \\
        &Att2in \cite{Rennie2016SelfCriticalST} &0.248 &0.134 &0.116 &0.091 &0.309 &0.162 \\
        &AdaAtt \cite{lu2017knowing}  &0.284 &0.207 &0.150 &0.126 &0.165 &0.311 \\
        &H\_RNN \cite{yin2019automatic} & 0.445& 0.292& 0.201& 0.154& 0.175& 0.344 \\
        & CMCL\cite{liu-etal-2021-competence} & 0.473 & 0.305 & 0.217 & 0.162 & 0.186 & 0.378 \\
        & R2Gen \cite{chen-etal-2020-generating} & 0.470 & 0.304 & 0.219 & 0.165 & 0.187 & 0.371 \\
        & R2Gen-CMN \cite{chen-etal-2021-cross-modal} & 0.475 & 0.309 & 0.222 & 0.170 & 0.191 & 0.375 \\
        & \textbf{Ours} & \textbf{0.492} & \textbf{0.318} & \textbf{0.230} & \textbf{0.175} & \textbf{0.199} & \textbf{0.381} \\
        \hline
        \hline
        \multirow{6}{*}{\textbf{MIMIC-CXR}} 
        &Show-tell \cite{xu2015show}& 0.243 &0.130 &0.108 &0.078 &0.307& 0.157 \\
        &Att2in \cite{Rennie2016SelfCriticalST}&0.314 &0.198 &0.133 &0.095 &0.122 &0.264 \\
        &AdaAtt \cite{lu2017knowing}&0.314 &0.198 &0.132 &0.094 &0.128 &0.267 \\
        
        &FVE \cite{nazarov2022importance} &0.299 &0.182 &0.124 &0.090 &0.123&0.238 \\
        & CMCL \cite{liu-etal-2021-competence} & 0.344 & 0.217 & 0.140 & 0.097 & 0.133 & 0.281 \\
        & R2Gen \cite{chen-etal-2020-generating} & 0.353 & 0.218 & 0.145 & 0.103 & 0.142 & 0.277 \\
        & R2Gen-CMN \cite{chen-etal-2021-cross-modal} & 0.353 & 0.218 & \textbf{0.148} & \textbf{0.106} & 0.142 & \textbf{0.278} \\
        & \textbf{Ours} &\textbf{0.360}&\textbf{0.218}&0.144&0.102&\textbf{0.144}&0.274 \\
        \hline
    \end{tabular}
    \label{tab:comparison}
\end{table*}

\subsection{Implementation details}
To maintain consistency with prior experimental setups \cite{chen-etal-2020-generating,li2018hybrid}, we utilize two images per patient as input for report generation on the IU X-RAY dataset, while for MIMIC-CXR, a single image is used. For visual feature extraction, we adopt ResNet101 to extract patch features, each encoded as a 2048-dimensional vector. For captioning generator, both Transformer encoder and decoder have a dimension of 512, with 3 layers, 8 attention heads, and initialized randomly. Anatomical attention alignment is designed using cross-attention with  8 attention heads, and a hidden state dimension of 512. Optimization is performed using the Adam optimizer, with an initial learning rate of \(1 \times 10^{-4}\) for the visual extractor and \(5 \times 10^{-4}\) for the remaining parameters, applying a decay rate of 0.8 per epoch. We train with a batch size of 16, running for 30 epochs on MIMIC-CXR and 100 epochs on IU X-RAY to achieve optimal performance. During inference, a beam size of 3 is used to balance generation quality and computational efficiency. 

\subsection{Compare with Previous Work}
\label{ssec:subhead}

Table~\ref{tab:comparison} shows the performance comparisons between previous studies and our proposed A3Net on the IU X-Ray and MIMIC-CXR datasets. A3Net consistently outperforms most prior methods, demonstrating strong results across both datasets. For the IU X-Ray dataset, A3Net achieves the highest BLEU-1 (0.492), BLEU-2 (0.318), and BLEU-3 (0.230) scores, surpassing R2Gen \cite{chen-etal-2020-generating} and CMCL \cite{liu-etal-2021-competence}. The model also outperforms R2Gen-CMN \cite{chen-etal-2021-cross-modal}, in BLEU-1 by 1.7\%, BLEU-2 by 0.9\%, and BLEU-3 by 0.8\%. Notably, our METEOR score of 0.199 outperforms R2Gen-CMN (0.191), showing an improvement of 4.2\%, indicating its superior ability to align with human-written reports in terms of lexical matching, word order, and morphological variations. Additionally, A3Net shows an improvement of 1.6\% in ROUGE-L (0.381) compared to R2Gen-CMN (0.375), further highlighting its ability to capture key reference text patterns. In the MIMIC-CXR dataset, A3Net achieves competitive BLEU scores (0.360 in BLEU-1, 0.218 in BLEU-2) and a strong METEOR score (0.144), outperforming other models. Although it shows slightly lower performance in BLEU-3 (0.144), BLEU-4 (0.102) and ROUGE-L (0.274) compared to R2Gen-CMN, it achieves a higher METEOR score (0.144 vs. 0.142), suggesting better alignment with human-written reports.

\begin{figure*}[]
    \centering
    \includegraphics[width=\textwidth]{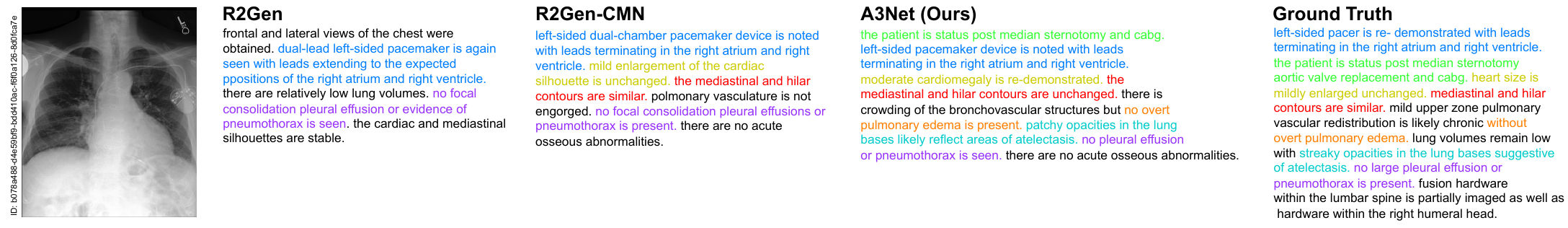} 
    \caption{ Qualitative results of our method. The text in bold color represents various diagnoses that appear in Ground Truth.}
    \label{fig:example}
\end{figure*}

We have selected an example image from the MIMIC-CXR test set for a case study of a report generated by different models. As shown in the Fig.~\ref{fig:example}, ground truth represents the actual reports annotated by radiologists, A3Net (Ours) refers to the report generated by our proposed method. Disease-related phrases in the ground truth are highlighted in different colors. It can be seen that A3Net generates more precise and informative reports, capturing key clinical findings more accurately than competing methods, demonstrating superior performance in both accuracy and clarity.
\subsection{Ablation study} 
To investigate the contribution of each visual feature in our approach, we conduct a rigorous ablation study on the IU X-Ray dataset. The results in Table~\ref{tab:ablation} demonstrate that both visual and semantic features play a crucial role in enhancing performance. The removal of any of them leads to a notable decline across all metrics, highlighting their individual prominent effectiveness. In particular, our proposed hyper-visual representation consistently improves performance across all metrics, reinforcing its effectiveness in bridging low-level visual perception and high-level semantic reasoning for superior report generation.
\begin{table}[t]
    \centering
    \footnotesize
    \renewcommand{\arraystretch}{1.2}
    \begin{tabular}{l ccc ccc}
        \toprule
        \textbf{Model} & \textbf{BL-1} & \textbf{BL-2} & \textbf{BL-3} & \textbf{BL-4} & \textbf{MTR} & \textbf{RG-L} \\
        
        \midrule
         Ours &  \textbf{0.492} & \textbf{0.318} & \textbf{0.230} & \textbf{0.175} & \textbf{0.199} & \textbf{0.381} \\
         $w/o \; \Phi_{\text{visual}}$ & 0.473 & 0.305 & 0.225&0.173&0.184&0.368 \\
         $w/o \; \Phi_{\text{sem}}$ & 0.475 & 0.309 & 0.222 & 0.170 & 0.191 & 0.375     \\
       
        \bottomrule
    \end{tabular}%

    \caption{The ablation study results on the IU X-Ray, with the best values highlighted in bold.}
    \label{tab:ablation}
\end{table}
\section{Conclusion}
\label{sec:majhead}In this work, we introduce A3Net, a radiology report generation model that incorporates an anatomical attention alignment mechanism to enhance vision-language understanding. By embedding structured knowledge of body part semantics into the feature space, A3Net improves perceptual awareness, enabling more precise associations between image regions and their corresponding anatomical structures. This hierarchical representation effectively bridges the gap between low-level visual features and high-level semantic reasoning, resulting in more accurate and contextually rich textual descriptions. Extensive experiments on the IU X-Ray and MIMIC-CXR datasets demonstrate that A3Net outperforms existing methods

\bibliographystyle{IEEEbib}
\bibliography{strings,refs}

\begin{thebibliography}{10}

\bibitem{xu2015show}
Kelvin Xu,
\newblock ``Show, attend and tell: Neural image caption generation with visual attention,''
\newblock {\em arXiv preprint arXiv:1502.03044}, 2015.

\bibitem{Rennie2016SelfCriticalST}
Steven~J. Rennie, Etienne Marcheret, Youssef Mroueh, Jerret Ross, and Vaibhava Goel,
\newblock ``Self-critical sequence training for image captioning,''
\newblock {\em 2017 IEEE Conference on Computer Vision and Pattern Recognition (CVPR)}, pp. 1179--1195, 2016.

\bibitem{lu2017knowing}
Jiasen Lu, Caiming Xiong, Devi Parikh, and Richard Socher,
\newblock ``Knowing when to look: Adaptive attention via a visual sentinel for image captioning,''
\newblock in {\em Proceedings of the IEEE conference on computer vision and pattern recognition}, 2017, pp. 375--383.

\bibitem{chen-etal-2020-generating}
Zhihong Chen, Yan Song, Tsung-Hui Chang, and Xiang Wan,
\newblock ``Generating radiology reports via memory-driven transformer,''
\newblock in {\em Proceedings of the 2020 Conference on Empirical Methods in Natural Language Processing (EMNLP)}, Bonnie Webber, Trevor Cohn, Yulan He, and Yang Liu, Eds., Online, Nov. 2020, pp. 1439--1449, Association for Computational Linguistics.

\bibitem{wang2022medical}
Zhanyu Wang, Mingkang Tang, Lei Wang, Xiu Li, and Luping Zhou,
\newblock ``A medical semantic-assisted transformer for radiographic report generation,''
\newblock in {\em International Conference on Medical Image Computing and Computer-Assisted Intervention}. Springer, 2022, pp. 655--664.

\bibitem{nazarov2022importance}
Otabek Nazarov, Mohammad Yaqub, and Karthik Nandakumar,
\newblock ``On the importance of image encoding in automated chest x-ray report generation,''
\newblock {\em arXiv preprint arXiv:2211.13465}, 2022.

\bibitem{yin2019automatic}
Changchang Yin, Buyue Qian, Jishang Wei, Xiaoyu Li, Xianli Zhang, Yang Li, and Qinghua Zheng,
\newblock ``Automatic generation of medical imaging diagnostic report with hierarchical recurrent neural network,''
\newblock in {\em 2019 IEEE international conference on data mining (ICDM)}. IEEE, 2019, pp. 728--737.

\bibitem{chen-etal-2021-cross-modal}
Zhihong Chen, Yaling Shen, Yan Song, and Xiang Wan,
\newblock ``Cross-modal memory networks for radiology report generation,''
\newblock in {\em Proceedings of the 59th Annual Meeting of the Association for Computational Linguistics and the 11th International Joint Conference on Natural Language Processing (Volume 1: Long Papers)}, Chengqing Zong, Fei Xia, Wenjie Li, and Roberto Navigli, Eds., Online, Aug. 2021, pp. 5904--5914, Association for Computational Linguistics.

\bibitem{liu-etal-2021-competence}
Fenglin Liu, Shen Ge, and Xian Wu,
\newblock ``Competence-based multimodal curriculum learning for medical report generation,''
\newblock in {\em Proceedings of the 59th Annual Meeting of the Association for Computational Linguistics and the 11th International Joint Conference on Natural Language Processing (Volume 1: Long Papers)}, Chengqing Zong, Fei Xia, Wenjie Li, and Roberto Navigli, Eds., Online, Aug. 2021, pp. 3001--3012, Association for Computational Linguistics.

\bibitem{sun2024r2gen}
Yongheng Sun, Yueh~Z Lee, Genevieve~A Woodard, Hongtu Zhu, Chunfeng Lian, and Mingxia Liu,
\newblock ``R2gen-mamba: A selective state space model for radiology report generation,''
\newblock {\em arXiv preprint arXiv:2410.18135}, 2024.

\bibitem{you-etal-2022-jpg}
Jingyi You, Dongyuan Li, Manabu Okumura, and Kenji Suzuki,
\newblock ``{JPG} - jointly learn to align: Automated disease prediction and radiology report generation,''
\newblock in {\em Proceedings of the 29th International Conference on Computational Linguistics}, Nicoletta Calzolari, Chu-Ren Huang, Hansaem Kim, James Pustejovsky, Leo Wanner, Key-Sun Choi, Pum-Mo Ryu, Hsin-Hsi Chen, Lucia Donatelli, Heng Ji, Sadao Kurohashi, Patrizia Paggio, Nianwen Xue, Seokhwan Kim, Younggyun Hahm, Zhong He, Tony~Kyungil Lee, Enrico Santus, Francis Bond, and Seung-Hoon Na, Eds., Gyeongju, Republic of Korea, Oct. 2022, pp. 5989--6001, International Committee on Computational Linguistics.

\bibitem{vaswani2017attention}
A~Vaswani,
\newblock ``Attention is all you need,''
\newblock {\em Advances in Neural Information Processing Systems}, 2017.

\bibitem{demner2016preparing}
Dina Demner-Fushman, Marc~D Kohli, Marc~B Rosenman, Sonya~E Shooshan, Laritza Rodriguez, Sameer Antani, George~R Thoma, and Clement~J McDonald,
\newblock ``Preparing a collection of radiology examinations for distribution and retrieval,''
\newblock {\em Journal of the American Medical Informatics Association}, vol. 23, no. 2, pp. 304--310, 2016.

\bibitem{johnson2019mimic}
Alistair~EW Johnson, Tom~J Pollard, Nathaniel~R Greenbaum, Matthew~P Lungren, Chih-ying Deng, Yifan Peng, Zhiyong Lu, Roger~G Mark, Seth~J Berkowitz, and Steven Horng,
\newblock ``Mimic-cxr-jpg, a large publicly available database of labeled chest radiographs,''
\newblock {\em arXiv preprint arXiv:1901.07042}, 2019.

\bibitem{li2018hybrid}
Yuan Li, Xiaodan Liang, Zhiting Hu, and Eric~P Xing,
\newblock ``Hybrid retrieval-generation reinforced agent for medical image report generation,''
\newblock {\em Advances in neural information processing systems}, vol. 31, 2018.

\end{thebibliography}

\end{document}